\documentclass{article}
\pdfoutput=1
\usepackage{tkai}
\usepackage{float} 
\usepackage[T1]{fontenc}
\usepackage[utf8]{inputenc}
\usepackage{times}  
\usepackage{helvet}  
\usepackage{courier}  
\usepackage[hyphens]{url}  
\usepackage{graphicx} 
\usepackage{natbib}  
\usepackage{caption} 
\usepackage{algorithm}
\usepackage{algorithmic}
\usepackage{enumitem}
\usepackage{booktabs}       
\usepackage{amsfonts}       
\usepackage{nicefrac}       
\usepackage{xcolor}         
\usepackage{amsmath}
\usepackage{amssymb}
\usepackage{color}
\usepackage{graphics}
\usepackage{lipsum}
\usepackage{multirow}
\usepackage{stmaryrd}

\usepackage{subcaption}
\newtheorem{theorem}{Theorem}[section]
\newtheorem{definition}[theorem]{Definition}

\newtheorem{corollary}[theorem]{Corollary}

\usepackage{breqn}

\usepackage{newfloat}
\usepackage{listings}
\DeclareCaptionStyle{ruled}{labelfont=normalfont,labelsep=colon,strut=off} 
\floatstyle{ruled}
\newfloat{listing}{tb}{lst}{}
\floatname{listing}{Listing}
\title{Stability Analysis of Various Symbolic Rule Extraction Methods \\ from Recurrent Neural Network}
\author{%
  Neisarg Dave \\
  The Pennsylvania State University\\
  \texttt{nud83@psu.edu}
  \And
  Daniel Kifer \\
  The Pennsylvania State University\\
  \texttt{duk17@psu.edu}
  \AND
  C. Lee Giles \\
  The Pennsylvania State University \\
  \texttt{clg20@psu.edu}  
  \And
  Ankur Mali \\
  University of South Florida\\
  \texttt{ankurarjunmali@usf.edu}
}

\begin{document}
\maketitle

\begin{abstract}
This paper analyzes two competing rule extraction methodologies: quantization and equivalence query. We trained $3600$ RNN models, extracting $18000$ DFA (Deterministic Finite Automata) with a quantization approach (k-means and SOM) and $3600$ DFA by equivalence query($L^{*}$) methods across $10$ initialization seeds. We sampled the datasets from  $7$ Tomita and $4$ Dyck grammars and trained them on $4$ RNN cells: LSTM, GRU, O2RNN, and MIRNN. The observations from our experiments establish the superior performance of O2RNN and quantization-based rule extraction over others. $L^{*}$, primarily proposed for regular grammars, performs similarly to quantization methods for Tomita languages when neural networks are perfectly trained. However, for partially trained RNNs, $L^{*}$ shows instability in the number of states in DFA, e.g., for Tomita 5 and Tomita 6 languages, $L^{*}$ produced more than $100$ states. In contrast, quantization methods result in rules with a number of states very close to ground truth DFA. Among RNN cells, O2RNN produces stable DFA consistently compared to other cells. For Dyck Languages, we observe that although GRU outperforms other RNNs in network performance, the DFA extracted by O2RNN has higher performance and better stability. The stability is computed as the standard deviation of accuracy on test sets on networks trained across $10$ seeds. On Dyck Languages, quantization methods outperformed $L^{*}$ with better stability in accuracy and the number of states. $L^{*}$ often showed instability in accuracy in the order of  $16\% - 22\%$ for GRU and MIRNN while deviation for quantization methods varied in $5\% - 15\%$. In many instances with LSTM and GRU,  DFA's extracted by $L^{*}$ even failed to beat chance accuracy ($50\%$), while those extracted by quantization method had standard deviation in the $7\%-17\%$ range. For O2RNN, both rule extraction methods had a deviation in the $0.5\% - 3\%$ range.
\end{abstract}

\section{Introduction}
Stateful networks such as recurrent networks (RNNs) effectively model complex temporal and noisy patterns on real-world sequences. In particular, RNNs with minimal syntactic grammatical errors can effectively recognize grammatical structure in sequences. This is evident by their ability to generate and generalize structure data, such as language \cite{bahdanau2015neural}, source code (C, C++, Latex, etc.) \cite{Tufano-Learning-CodeChanges} \cite{Tufano_bug_fixes}, and program synthesis \cite{zhang_program_synthesis}. To better understand the internal workings of RNNs, recent works are focused on extracting interpretable rules from trained RNNs that better evaluate the ability of RNNs to recognize grammatical structures \cite{Zeng_dfa_clustering_rnn} \cite{Wang2017AnEE} \cite{weiss2018extracting}. However, the stability of these rule extraction needs to be better studied, as instability often leads to failure in extraction. This work aims to explain the internal workings of various types of RNNs. We analyze their stability through rigorous comparison to fundamental concepts in formal languages, i.e., automata theory.

Recent theoretical works have shown that with some constraints, such as finite precision and time, RNNs and GRU can only recognize regular grammars \cite{Merrill_2019}. However, another class of RNNs, 02RNN, has a much more robust construction and has shown equivalence with only $N+1$ state neurons for DFA with $N$ states and is argued to be stable \cite{mali2023computational, STOGIN2024120034}. This paper thoroughly analyzes the stability of the internal state of various types of RNNs, such as first-order (RNN, GRU, LSTM) and second-order RNN (O2RNN). 
One family of approaches widely used in literature to understand the internal workings of RNNs is to extract an interpretable state machine from the trained RNN. Two widely used approaches that take inspiration from grammatical induction are L-star \cite{angluin1987learning, weiss2018extracting} and Quantization-based approaches \cite{giles1993extraction, omlin1996extraction}. L-star is an active learning approach that learns via two forms of queries. The first is the membership query, which focuses on strings in language L, and the other is the equivalence query, which compares a candidate DFA to the true or oracle DFA. However, L-star requires RNNs to be 100\% accurate, which is problematic in a real-world scenario as NNs rarely get achieve 100\%. Second, they also assume these black box models that compute oracles to be available at train time, which is again problematic for large, expensive equivalence queries. Other works are based on the quantization approach, which uses clustering (k-means, SOM) on the RNN hidden states to extract the state machines and does not suffer from the above problem. However, none of the prior work analyzes the stability of these rule extraction methods, as NNs are stochastic and introduces various forms of uncertainty resulting in generalization issue. Thus, this also hampers the extraction strength of various extraction approaches, resulting in sub-optimal state machines.

\paragraph{This Work}We conduct an extensive empirical study to discover a better stable extraction approach across a wide variety of RNNs. This work also sheds light on whether the theoretical limitations of RNNs hold in practice when trained using backpropagation of error through time. We make the following main contributions:
\begin{enumerate}
    \item We show that quantization-based rule extraction is more stable than the equivalence query-based method.
    \item Quantization-based approach performs better, especially when RNNs cannot reach $100\%$ accuracy.
    \item Quantization-based approach performs better when underlying grammar contains complex patterns.
    \item We empirically show that $2^{nd}$ RNNs encapsulate stable rules than $1^{st}$ order networks. Thus empirically validating prior theoretical work.
    \item We conduct extensive experiments across different RNN architectures with varied hidden state sizes and grammar types to show the stability of the rule extraction method.
\end{enumerate}

\section{Related Work}
RNNs are known to be excellent at recognizing patterns in text and algorithmic patterns. Extensive work is focused on deriving computational power and expressivity of these systems. For instance, memory-less RNNs with infinite precision are known to be Turing complete \cite{Siegelmann_1995, mali2023computational}, whereas memory-augmented RNNs are Turing Complete even with finite precision \cite{STOGIN2024120034, mali2023tensor}. \cite{funahashi1993approximation} showed there exists a class of RNN that can approximately represent dynamic systems, whereas \cite{rabusseau2019connecting} showed that linear second-order RNN is equivalent to weighted automaton.

Recently \cite{Merrill_2020} classified NNs based on space complexity and rational recurrence. They show that GRU and RNN can model regular grammars with $O(1)$ while LSTM can also model counter machines with space complexity of $O( \log n)$ where $n$ is the input sequence length. \cite{Hewitt_2020} show that the theoretical bound of $O(m \log k)$ for a Dyck-($k$, $m$) language on the size of hidden units is too tight and is not achievable even by unbounded computation. Dyck-($k$, $m$) languages have $k$ types of nested parenthesis with a maximum depth of $m$ nests. The best empirical results are obtained at O($k^{\frac{m}{2}})$.

Recent works are more focused on exploring the relationship between RNN internals and DFAs through a variety of methods. One such direction is based on automata extraction and grammatical induction. \cite{angluin1987learning} proposed an equivalence query-based approach, $L^*$,  to construct DFA (Deterministic Finite Automata) from a set of given examples. The method requires an oracle to classify positive and negative examples also provide counterexamples. \cite{weiss2018extracting} used RNN models as oracles to $L^*$ algorithm. \cite{giles1993extraction} and \cite{omlin1996extraction} showed quantization-based extraction mechanisms for symbolic rules from Recurrent Neural Networks in the form of DFAs. This was further supported by \cite{Wang2017AnEE} with empirical results. All these prior results primarily focus on Tomita Grammars/Languages \cite{tomita1982learning}. Tomita languages have since been used to demonstrate the computational capacity of neural networks for regular grammars.

\cite{sennhauser-berwick-2018-evaluating,suzgun2019lstm, mali2021investigating} provided the empirical evaluation of LSTM on Context-Free Grammars. \cite{suzgun2019lstm, mali2021investigating} show that LSTM with just one hidden unit is sufficient to learn Dyck-1 grammar. \cite{Bhattamishra_2020} experimented with self-attention models to demonstrate their ability to learn counter languages like $n$-ary Boolean Expressions. \cite{Ebrahimi_2020} showed that networks with self-attention perform better than LSTMs in recognizing Dyck-$n$ languages.  \cite{Suzgun_2019, mali2021recognizing} trained memory augmented RNNS to learn Dyck Languages with up to 6 parenthesis pairs. \cite{Mali_2020} improved upon the memory-augmented Tensor RNNs and used an external differentiable stack to recognize long strings of Context-Free Grammars. Recently \cite{deletang2022neural, mali2020neural} conducted the most comprehensive study, which places several types of NNs based on the Chomsky hierarchy. However, \cite{deletang2022neural} work was primarily focused on first-order NNs. 

Prior works have largely focused on the idea of encoding and learnability of regular and context-free grammars by Recurrent Neural Networks and transformers. Theoretical bounds, empirical bounds, and rule extraction methods remain the main focus of research. The question of the empirical stability of these models and the rules extracted from them have largely gone unanswered. As often the theoretical guarantee doesn't provide any polynomial time convergence or stability guarantee. In this paper we try to empirically validate these claims and focus on empirical stability, we perform large-scale experiments and report our findings on the stability of RNNs and the two rule extraction methods (quantization and equivalence query) on Tomita and Dyck languages.

\section{Background \& Methodology}

\paragraph{Recurrent Neural Networks}
To evaluate the stability of extraction methods, we train four types of RNN models. We use LSTM \cite{Hochreiter1997LongSM} and GRU \cite{DBLP:journals/corr/ChungGCB14} for $1^{st}$ order networks and O2RNN \cite{giles1989higher} and MIRNN \cite{DBLP:journals/corr/WuZZBS16} as $2^{nd}$ order networks. We briefly describe the state update equation in O2RNN which has a higher order operation facilitated by a $2^{nd}$ order weight tensor given as follows:
\begin{align*}
    h_i^{(t+1)} = \sigma(\sum_{j, k} W_{ijk}x_i^{(t)}h_k^{(t)} + b_i)
\end{align*},
where $h$ is the state transition of the O2RNN such that $h_t \in \mathbb{Q}$ and $\mathbb{Q}$ is a set of rational numbers, $W \in \mathbb{R}^{n \times n \times n}$ contains the $2^{nd}$ order recurrent weight matrices, $b \in \mathbb{R}^n$ are the biases, $sigma$ is the nonlinear activation function such as the logistic sigmoid, and $x$ is the input symbol at time $t$. It is theoretically shown that 02RNN can simulate DFA with finite precision and time with only $n+1$ state neurons.
\begin{theorem} \cite{mali2023computational}
    Given a DFA $M$ with $n$ states and $m$ input symbols, there exists a $k$-neuron bounded precision O2RNN with sigmoid activation function ($h_H$), where $k = n+1$, initialized from an arbitrary distribution, that can simulate any DFA in real-time O($T$).
\end{theorem}
Furthermore, it was proved that the above construction using O2RNN is stable. This is important in our study as we empirically try to show the stability when models are trained using gradient-based learning.
To validate above hypotheses we also experiment with noisy second-order connection that provides faster inference, the network known as MIRNN which approximates the second-order connection by introducing rank-1 approximation in the form of multiplicative integration \cite{DBLP:journals/corr/WuZZBS16}. This is achieved by replacing multiplication with Hadamard product in the vanilla RNN equation. Thus the state transition or update equation for MIRNN is written as follows:
\begin{align*}
    h^{(t+1)} = \sigma( \alpha Ux^{(t)} \odot Vh^{(t)} + \beta_1 U \odot x^{(t)} + \beta_2V \odot h^{(t)})
\end{align*} \footnote{Bias omitted only for representation, but network has standard Bias term}
where $\alpha$, $\beta_1$ and $\beta_2$ are trainable parameters or can be considered as additional sets of bias vectors, and $\odot$ denotes Hadamard product. 

\paragraph{Deterministic finite automaton}
A Deterministic Finite Automaton is a $5$-tuple ($Q, \Sigma, \delta, q_0, F$). $\Sigma$ is the set of symbols called alphabet which are used to create strings. A DFA can either accept or reject an input string. A language recognized by DFA is a set of strings accepted by the DFA. Grammar is a set of rules followed by the language. DFA acts as a machine that checks if a string follows the grammatical rules. Hence, DFA are considered to be the representation of grammatical rules. 

$Q$ is the set of all states in the DFA. $q_0 \in Q$ is the initial state of the DFA and $F \subseteq Q$ is the set of  final accepting states. Any state $q \in Q$ s.t. $q \notin F$ is a rejecting state. The transition matrix $ \delta: Q \times \Sigma \rightarrow Q$ encodes the rules of state transitions.

\paragraph{DFA Extraction}
%

There are two prominent techniques for DFA extraction from RNNs: quantization-based \cite{omlin1996extraction}\cite{giles1993extraction} and counter-example based \cite{angluin1987learning}\cite{weiss2018extracting}.

In a quantization-based approach, hidden states {\color{black}($h_{ij}$)} from each step {\color{black}($j$)} of RNN computation for sample {\color{black}($i$)} are collected and clustered into {\color{black}$k$} groups {\color{black}$G = \{g_1, g_2, ... g_k\}$}. {\color{black} Here we can choose a clustering algorithm $C$ such that $C(h_{ij}) = g_m \in G$}. These groups form one-to-one correspondence with the states of the DFA {\color{black}($Q = G$)}. A transition matrix {\color{black}($\delta(C(h_{ij}), a_{ij}) \rightarrow g_n \in G$)} is then created by mapping group transitions for each input {\color{black} $a_{ij} \in \Sigma$}. Initial state {\color{black} $q_0 = C(h_{init})$} and final states {\color{black}($F$)} are selected through majority voting of qualified inputs. This gives us an initial DFA, which is then minimized \cite{hopcroft1971n} to get a final minimal DFA. In this work, we use k-means \cite{lloyd1982least} and self-organizing maps (SOM) \cite{kohonen1982self} for clustering. For each model we extract $5$ DFA's with a successively increasing number of states. These DFA's are minimized, and the best performing DFA (on validation set) is selected.  With $3200$ models and $5$ extraction per model, we extract $16000$ DFAs.

$L^*$ proposed by \cite{angluin1987learning} creates a transition function of states by repeatedly asking an oracle to classify strings. The oracle is also responsible for providing counter-examples. {\color{black} These concepts are explained in depth in Appendix \ref{app:lstar}}. \cite{weiss2018extracting} use RNN models as oracle for $L^*$. We follow their approach to extract DFA using $L^*$ for each of the $3200$ models. One observation we made here is that the extraction heavily depends on initial counter-examples. This does not prove good for Dyck grammars especially.

\paragraph{Language from a Concentric Ring Representation}

\begin{figure*}[h]
\includegraphics[width=0.8\textwidth]{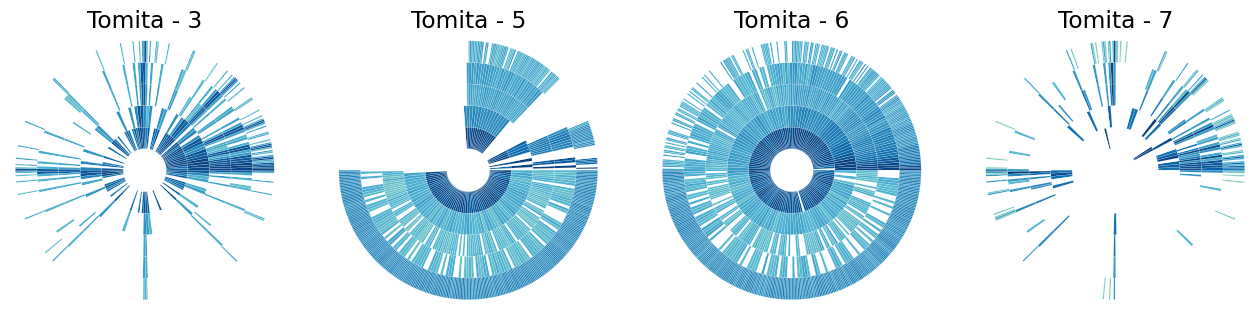}
\centering
\caption{Distribution of strings of Tomita grammar in lexicographical order }
\label{fig:tomita_dist}
\end{figure*}

\begin{figure*}[h]
\includegraphics[width=0.8\textwidth]{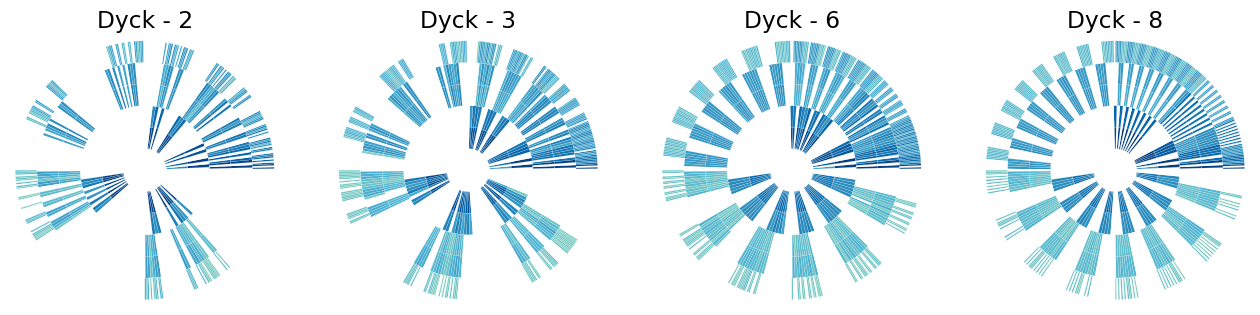}
\centering
\caption{Distribution of strings of Dyck grammar in lexicographical order}
\label{fig:dyck_dist}
\end{figure*}

The concentric ring representation is used to analyze the complexity of a language \cite{watrous1991induction}. It reflects the distribution of its associated positive and negative strings within a certain length. In each concentric ring, we can observe that all strings with the same length are arranged in a lexicographic order where white represents positive string and black/blue represents negative strings. The more variation one observes in a concentric ring, the more difficulty neural networks have in learning. In figure \ref{fig:tomita_dist} and \ref{fig:dyck_dist} we show the ring presentation for Tomita and dyck grammar, respectively. 

\color{black}
\paragraph{Stability Analysis}
\begin{theorem}
    BROUWER’S FIXED POINT THEOREM \cite{boothby-fpt}: For any continuous mapping $f:Z \rightarrow Z$, where Z is a compact, non-empty convex set, $\exists \ z_f$ s.t. $f(z_f) \rightarrow z_f$
    \label{th:fpt}
\end{theorem}

\begin{definition}
Let $\mathcal{E}$\ be an estimator of fixed point $z_f$ for mapping $f:Z \rightarrow Z$, where $Z$ is a metric space. $\mathcal{E}$ is a stable estimator iff estimated fixed point $\hat z_f$ is in immediate vicinity of $z_f$, i.e. $ | \hat z_f - z_f | < \epsilon,\ $ for arbitrarily small $\epsilon$.
\end{definition}

 \begin{corollary} Stability of estimator $\mathcal{E}$\ can be shown by iteratively computing $z^{(t+1)} = \mathcal{E}_f(z^{(t)})$ \ with $z^{(0)}$ in the neighborhood of $z_f$. For stable estimator $\mathcal{E}$,\ \ $\lim_{t\rightarrow \infty}z^{(t)}  = z_f$. Neighborhood of $z_f$, $N_{z_f}$ is a set of points $z_n \in N_{z_f}$ near $z_f$ s.t. $|z_n - z_f| < \delta $.  
\end{corollary}

\cite{Omlin_Giles_sigmoid} use this idea of stability to show that for provably stable second-order RNNs, the sigmoidal discriminant function should drive neuron values to near saturation. Similarly, \cite{stogin2020provably} use the fixed point analysis to prove the stability of differentiable stack and differentiable tape with RNNs.

\begin{definition}
Let $\mathcal{E}^*$ be a stochastic fixed point estimator for $f$. The estimated $RMS$ error for fixed point estimation, calculated iteratively as $\hat z_{f}^{(i)} = \mathcal{E}_f^*(\hat z_{f}^{(i-1)})$ can be defined as:
$$
    e_{rms} = \sqrt{\frac{1}{n}\sum_{i=0}^{n} || \hat z_{f}^{(i)} - z_f ||^2}
$$
\label{def:rms}
Note that when true $z_f$ is unknown, the law of large numbers allows us to use the estimated mean $\overline{z}_f$ instead. Also the error estimate $e_{rms}$ is the standard deviation from $z_f$
\end{definition}

    
\textbf{Stability of States}: Let states $Q$ of DFA ($\Sigma, Q, \delta, q_0, F$) be embedded in the metric space of $\mathcal{R}^n$ and mapping $\delta : Q \times \Sigma \rightarrow Q$ be the transition function. Consider transitions for state $q \in Q$ s.t. $\delta(q, a) = q$ for some $a \in \Sigma$. By Theorem \ref{th:fpt} and Definition \ref{def:rms}, the error in stability of states can be estimated. The stability of states directly influences the stability of predictions.

{\color{black}
Thus, we wish to ask the following questions

\textbf{Research Question:}
\begin{enumerate}
    \item Does there exist a class of stable RNNs, that, for any regular language over dataset or distribution generated over any probability distribution ($\mathbb{P}_d$) clusters the internal states in reasonable time to approximate minimal DFA with high probability ($\mathbb{P}_m$), such that difference between states of RNN ($\hat{Q}$) and DFA ($Q$) in euclidean space is within the small bound.
$||Q - \hat{Q}||_2^2 < \epsilon$
    
    \item RNNs that are partially trained and do not achieve perfect accuracy (i.e. $100 \%$) are known as partially trained RNNs. We consider them as a stochastic state estimator for that language. How do different DFA extraction methods affect optimal state estimation from partially trained RNN such that the extracted state ($\hat{Q}_e$) is much closer to the optimal state ($\hat{Q}$) of stable RNN.
    $||\hat{Q} - \hat{Q}_e||_2^2 < \epsilon_e$, where $\epsilon_e \geq \epsilon$

\end{enumerate}
}
\color{black}


\section{Data and Models}
In all our experiments, RNNs (both first and second order) are trained on a dataset of positive and negative input example strings. These strings are 
randomly generated with varied lengths for any given formal language. Below, we explain our dataset creation procedure and training strategies.  

\textbf{Tomita Grammars}: One of the widely used datasets for benchmarking grammatical inference algorithms is based on 7 Tomita grammars (definitions are given in Table \ref{tab:tomita}) \cite{tomita1982learning}.   In this work, to have a fair comparison with the literature, we adopt the dataset creation script provided by \cite{bhattamishra2020ability}. However \cite{bhattamishra2020ability} do not provide \textbf{separate validation set} for model selection and hyperparameter optimization. This approach of optimizing models on a test set creates biased prediction. Thus to overcome this issue we use their bin\_0 as the validation set and generated our own test bin\_0 of $2000$ samples for each of the seven Tomita grammars. The train, validation, and test bin\_0 sets have a minimum string length of $2$ and a maximum $50$. We also create another test termed as test bin\_1 which has strings of length $\in [51, 100]$, this set test the extrapolation or generalization capability of the model. The train sets have $10000$ samples each, while validation and test sets have $2000$ samples each. To avoid any potential data leakage, we ensured that there was no overlapping of samples in any sets.

\textbf{Dyck Grammars}: The second dataset we use comes from a context free languages family, known as Dyck grammar. A Dyck word is a set of balanced parenthesis. A Dyck grammar can be written as a context-free grammar with production rule $S \rightarrow "(" S ")" | \epsilon $ where $"("$ and $")"$ are left and right parenthesis respectively and $S$ is a non-terminal. Note that the string generated from such a rule is well-balanced. 

Let $s_1, s_2 \in D$ be strings of dyck language $D$, then $s_1 + s_2 \in D$, where $+$ is a string concatenation operator. A string made up of $n$ pairs of well-nested parenthesis belongs to the Dyck-$n$ language.

We follow a split similar to Tomita datasets for train, validation, and test sets for Dyck-$2$,$3$,$6$, and $8$. For training sets, we generate $20,000$ samples for each dyck language, while for test and validation sets, we generate $4000$ samples each. Tables \ref{tab:num_samples} encapsulate the details.

\textbf{Partial Training:} The next setup we experimented is known is partial trained system, where model never reaches $100\%$ accuracy when tested on unseen distribution such as validation set. Often RNNs when trained on grammatical inference dataset achieve perfect accuracy (i.e $100\%$) on validation split. However, in the real world scenario and even while dealing with natural language, it is difficult for NNs to achieve perfect accuracy.  \textit{We argue that an ideal extraction approach should even function in scenarios where the model doesn't achieve the perfect score.} Thus, to simulate these settings and evaluate the stability of DFA extraction from a reasonably trained network, we create a set of partially trained neural networks on Tomita grammars. We stop the training of these networks as the validation accuracy reaches a threshold (i.e $>85\%$). However, some grammars like Tomita $1$ and $2$ often see higher accuracy within first few epochs and the  validation accuracy jumps around $69\%$ to $90\%$ between consecutive iterations. However, no network in this set achieves more than $90\%$ accuracy on validation and test sets. We provide complete training details as well as range of hyperparameters in appendix \ref{sec:appendix_training}


\section{Results and Discussion}
In this section, we will analyze the stability of DFA extraction of various grammars, respectively. Additional results are provided in appendix \ref{sec:add_result}.
\textbf{Tomita Grammars}

As discussed earlier, we trained four different RNNs on 7 Tomita grammar.  In this section, we will primarily compare results from Tomita 3, 5, 6, and 7. As based on concentric ring representation, these grammars are slightly more complex compared to others. In the appendix \ref{sec:add_result}, we show model performances on Tomita 1, 2, and 4, respectively. 



\paragraph{Performance of RNNs on Tomita Grammars}

In our experiments, we train networks with state dimensions of size $n.s$, where $s$ is a multiplier that we vary in $[1, 2, 3, 4, 5]$ and $n$ is the minimal number of states in the ground truth DFA of the grammar, as given in the seminal work \cite{tomita1982learning}. Tomita $6$ grammar has a minimal state dimension of $3$ (table \ref{tab:tomita_states}). From the concentric ring representation and analyzing the lexicographic space of grammar 6 (figure \ref{fig:tomita_dist}), we can observe grammar 6 has an evenly distributed pair of positive and negative strings. Thus, making Tomita 6 a fairly complicated regular grammar, prior works have empirically noticed similar behavior \cite{weiss2018extracting, Wang_2019}.  Thus, this provides us with an optimal way to examine the representation capability of  $1^{st}$ and $2^{nd}$ order RNNs. Ideally if model has better encoding capability then it should remain stable, despite the complexity of training distribution. This phenomenon is evident from figure \ref{fig:tomita_rnn_mean} where we can see that O2RNN consistently performs better than other RNNs representing higher learning capacity of $2^{nd}$ order recurrent cell operations. This is interesting to observe as \cite{Merrill_2019} has theoretically shown that LSTM is strictly more powerful than regular languages. 
We empirically validate the theoretical results in O2RNN that argue stability \cite{mali2023computational}, as it is evident from figure \ref{fig:tomita_rnn_mean}, which shows the mean accuracy of RNNs across $10$ seeds, where O2RNN constantly achieves higher accuracy across state-dimension.  In figure \ref{fig:tomita_rnn_std}, we also report the standard deviation of all the models across seeds. This is an important measure to provide statistical significance of any model and whether model can indeed learn the structural knowledge from the data. For simplicity, we show the standard deviation of models with a state dimension of $2n$, where $n$ is the number of states in the minimal ground truth DFA. However, these results do extrapolate to other state dimensions, and we observe consistent trends across settings. 

As we have observed above, with a special emphasis on Tomita-6, \textbf{\textit{O2RNN is more stable}} as it gives $\textbf{0\%}$ standard deviation compared to $\textbf{15.81\%}$, $\textbf{23.85\%}$ and $\textbf{26.35\%}$ deviation of LSTM, GRU and MIRNN respectively across $10$ seeds on test bin $0$. Similar performance was observed on test bin $1$. Given that O2RNN can insert \cite{giles1993extraction} and refine rules \cite{giles1992learning, mali2020neural} and often learn with fewer samples \cite{mali2023computational}, we believe these results are an important direction towards building neuro-symbolic AI systems.

\begin{figure*}[!htb]
\includegraphics[width=0.8\textwidth]{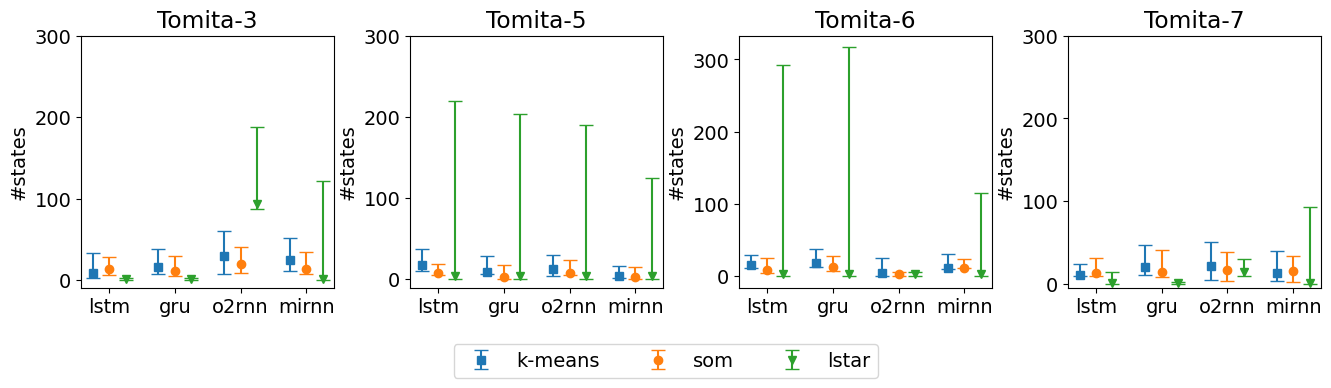}
\centering
\caption{Min, max, and mode of the number of states extracted from $1^{st}$ and $2^{nd}$ order RNNs trained on Tomita grammars}
\label{fig:tomita_dfa_states_comparison}
\end{figure*}

\textbf{Comparison of DFA extracted from Tomita Grammars:}
DFAs extracted using quantization and equivalence query methods show similar accuracies on Tomita languages. In figure \ref{fig:tomita_dfa_accuracy} we report the mean performance of DFA's extracted by quantization approach using self-organizing maps across $10$ seeds. Similarly, figure \ref{fig:tomita_lstar_accuracy} shows the mean performance of DFAs extracted using $L^{*}$. Figure \ref{fig:tomita_dfa_accuracy_comparison}  shows the standard deviation of performance of DFA's from mean accuracy across $10$ seeds. Here, we can observe that both methods show similar stability in terms of the performance of DFAs on the test set (bin-$0$). 
Even though the performance of DFAs extracted by the two methods is similar for Tomita languages, there is a noticeable difference in the number of states in the extracted DFAs. From figure \ref{fig:tomita_dfa_states_comparison} we can observe that the deviation in the number of states extracted from the mode number of states across $10$ seeds is quite large for DFAs extracted using $L^{*}$ . \textit{For example, we can observe from fig \ref{fig:tomita_dfa_states_comparison}, in Tomita-5, the maximum number of states extracted by \textbf{$L^{*}$ is above $\textbf{200}$} for LSTM, GRU, and O2RNN and about $150$ for MIRNN.}
Note that the number of states in Tomita-5 ground truth DFA is only $4$. However, the number of states extracted by quantization methods consistently remains low ( \textbf{under $\textbf{10}$ states} for the majority of seeds ). 

\begin{figure*}[htb!]
\includegraphics[width=0.8\textwidth]{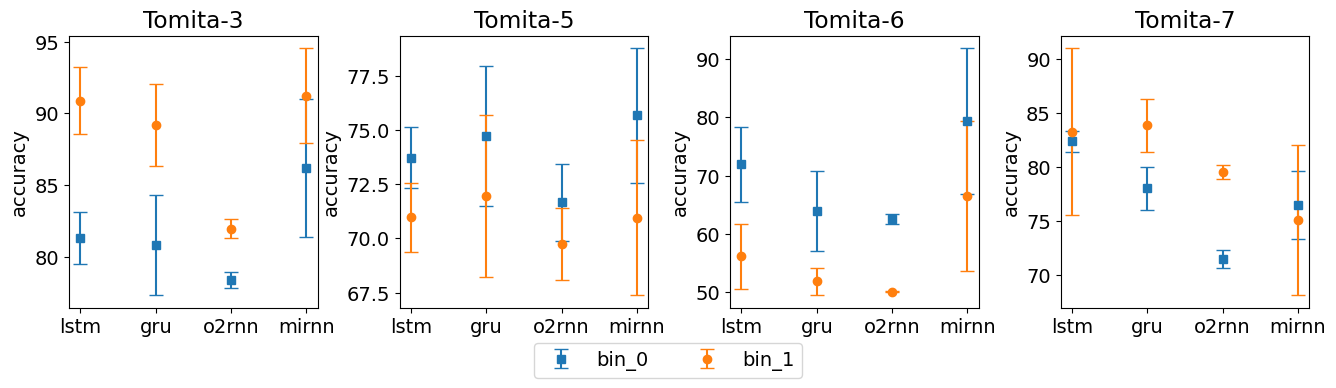}
\centering
\caption{Mean Accuracy and Standard Deviation of $1^{st}$ and $2^{nd}$ order Recurrent Neural Networks with restricted training on Tomita languages. The training was stopped when validation accuracy crossed $85\%$.}
\label{fig:tomita_rnn_std_0.85}
\end{figure*}
\begin{figure*}[h]
\includegraphics[width=0.8\textwidth]{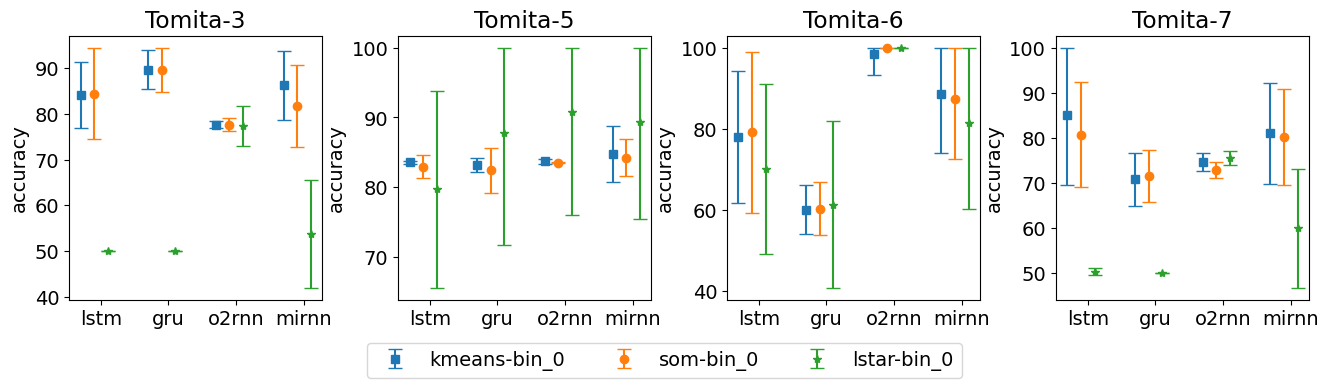}
\centering
\caption{Mean and Standard Deviation of  the accuracy of DFAs extracted from $1^{st}$ and $2^{nd}$ order RNNs with restricted training ($85\%$ validation accuracy) on Tomita languages.}

\label{fig:tomita_accuracy_dfa_0.85}
\end{figure*}


\textbf{Partially Trained Tomita Grammars}

As argued in the main paper, the DFA extraction from partially trained RNNs might provide a better insight into the stability of the two extraction methods. Furthermore it can also provide an insight whether a model can indeed learn the structural knowledge from the data. The Figure \ref{fig:tomita_rnn_std_0.85} shows the performance of partially trained RNNs on Test bin\_{0} and generalization set bin\_{1}, and figure \ref{fig:tomita_accuracy_dfa_0.85} shows the performance of  DFAs extracted from partially trained RNNs. Here, we can observe that $L^{*}$ fails to extract valid DFA for LSTM and GRU for Tomita-3 and Tomita-7 and also shows high instability for Tomita-5 and Tomita-6. In contrast, K-means and SOM-based extraction produces relatively stable DFAs. Also, we can observe that DFAs extracted from O2RNN by quantization methods are more stable as they show low deviation and better mean performance across $10$ seeds. Note that in Tomita-3 and Tomita-7, even though the mean performance of DFA extracted from LSTM is higher, it also has higher instability compared to O2RNN. \textit{In Tomita-6, which is a relatively difficult grammar to learn, DFA extracted from O2RNN has the highest mean accuracy and the lowest deviation.} \textbf{It is important to note that none of the prior work conducted this type of study with partially trained models.} From the result, it is evident that even in scenarios where the model exhibits lower generalization performance, quantization-based methods can still extract stable rules, which is an important directional towards building responsible AI systems.


\subsection{Dyck Languages} In this section, we analyze the stability of DFA extraction methods on dyck language.

\textbf{Performance of RNNs on Dyck Languages}

For all RNN variants, figure \ref{fig:dyck_rnn_mean} shows that the mean accuracy increases with an increase in state dimension before stabilizing, indicating that for complex grammars, more parameters are required. We observe that LSTM and GRU can easily reach near $100 \%$ accuracy with a lesser number of neurons in the hidden state compared to O2RNN and MIRNN. Figure \ref{fig:dyck_rnn_std} shows the standard deviation of the accuracy of RNNs on Dyck languages across $10$ seed. Similar to Tomita, we evaluate all models with hidden state dimension $=2k$ where $k$ is the type of Dyck language, i.e. Dyck-$k$ . We observe all networks behave similarly, whereas GRU shows highly better performance. However, GRU and LSTM are more sensitive to string lengths as the standard deviation is higher for test bin (generalization set) $1$, which has strings of length $51$-$100$ as compared to test bin $0$ with string lengths $2$-$50$. O2RNN and MIRNN show similar standard deviations for both test sets. \textit{Thus indicating higher-order networks better understand the structural property of the language as opposed to memorizing the input/train distribution.}


\textbf{Performance comparison of DFA extracted from Dyck Grammars}

\begin{figure*}[!tb]
\includegraphics[width=0.8\textwidth]{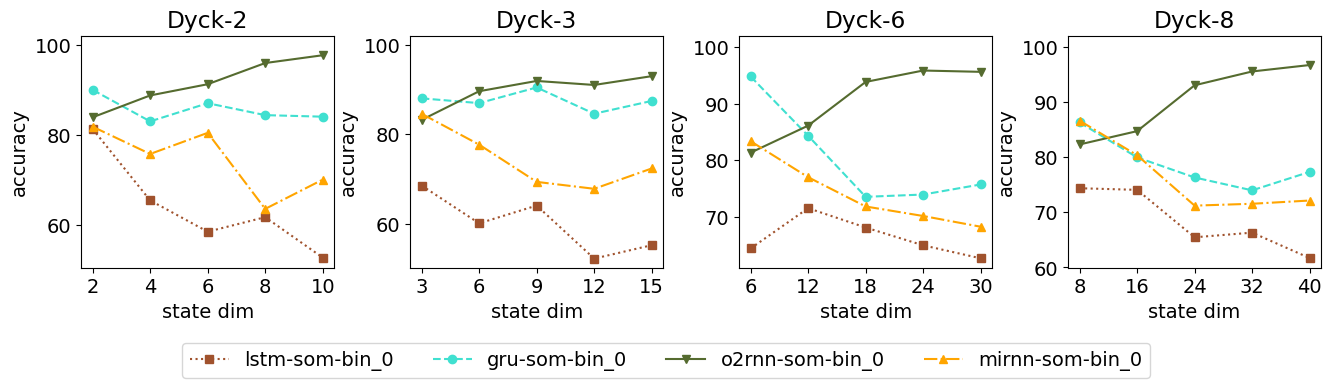}
\centering
\caption{Mean accuracy of DFA extracted from RNNs by quantization methods across $10$ seeds on Dyck grammars }
\label{fig:dyck_states_accuracy}
\end{figure*}

\begin{figure*}[!tb]
\includegraphics[width=0.8\textwidth]{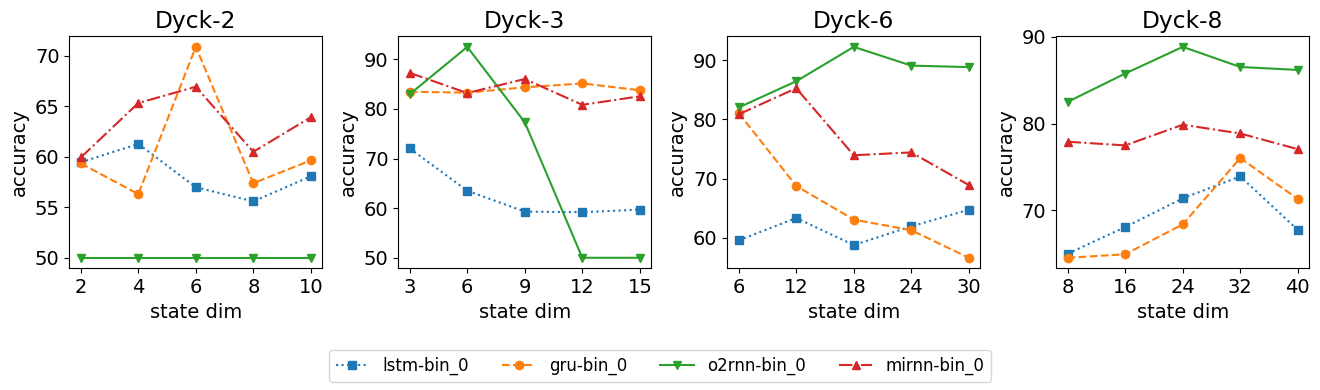}
\centering
\caption{Mean accuracy of DFA extracted from RNNs by $L^{*}$ across $10$ seeds on Dyck grammars}
\label{fig:dyck_lstar_accuracy}
\end{figure*}

\begin{figure*}[!tb]
\includegraphics[width=0.8\textwidth]{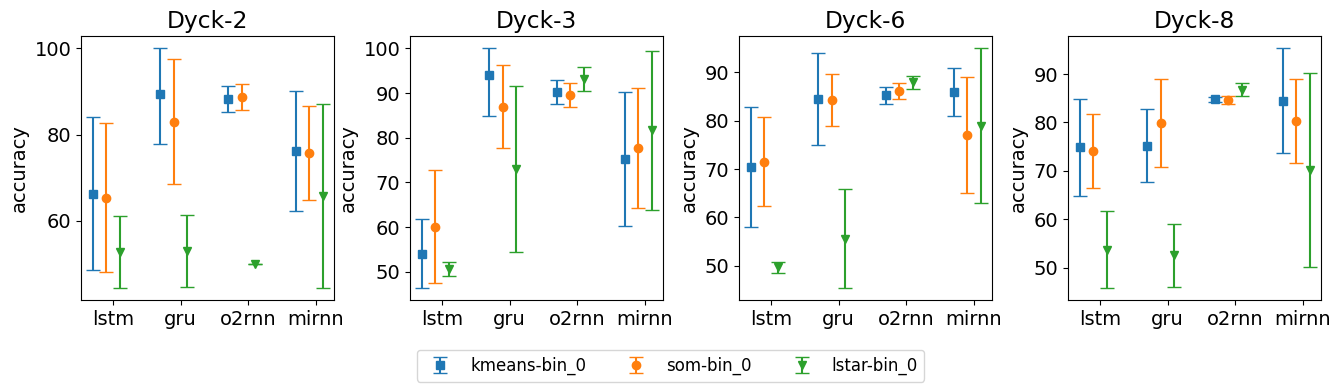}
\centering
\caption{Mean and Standard deviation of accuracy of DFA extracted from $1^{st}$ and $2^{nd}$ order RNNs on Dyck grammars.}

\label{fig:dyck_comparison_accuracy}
\end{figure*}

As highlighted earlier, GRU gets higher accuracy compared to other RNNs in recognizing Dyck languages. However, from figure \ref{fig:dyck_states_accuracy}, we can see that DFA extracted from O2RNN using the quantization method consistently outperforms other models DFAs over a broader range of hidden state dimensions. We can also see this phenomenon for DFAs extracted by $L^{*}$  (fig \ref{fig:dyck_lstar_accuracy}) for Dyck-$6$ and Dyck-$8$ languages too.  Figures \ref{fig:dyck_states_accuracy} and  \ref{fig:dyck_comparison_accuracy} show the ability of O2RNN to understand and encode rules most stably, which can be extracted effectively by using quantization methods. \textit{This shows that O2RNN is trying to learn the underlying grammatical structure as opposed to memorizing the language.}

Furthermore, by comparing figures \ref{fig:dyck_states_accuracy} and \ref{fig:dyck_lstar_accuracy}, we observe that DFAs extracted using quantization methods consistently outperform DFAs extracted using $L^{*}$.  This issue becomes even more evident by analyzing Figure \ref{fig:dyck_comparison_accuracy}, which shows the stability issues of $L^{*}$ based approaches while extracting DFA from Dyck languages. On multiple occasions, DFAs extracted by $L^{*}$ even fail to perform better than \textit{random guess} ($50\%$).  
We can also observe the instability of $L^{*}$ in the number of states in the extracted DFAs as we vary the state dimensions(figure \ref{fig:dyck_comparison_states_mode}) and even across $10$ seeds ( figure \ref{fig:dyck_comparison_states_std}.  The number of states in  DFA's extracted by $L^{*}$ can range in the order of $\textbf{100}$ states while quantization methods consistently produce DFA's with less than $\textbf{50}$ states.

\section*{Conclusion}
In this work, we studied how various classes of RNNs ($3600$ models) trained to recognize formal languages stably represent knowledge in their hidden state. Specifically, we have asked how stable and close these internal representations are compared to Oracle state machines (\textbf{R1}) and whether we can stably extract minimal deterministic state machines across various settings, even in scenarios where the network is partially trained (\textbf{R2}). To prove our hypothesis, we conduct extensive experiments and show that the state machine extracted from the $2^{nd}$ RNN network (O2RNN) is much more stable and achieves $0\%$ standard deviation, followed by LSTM with $4\%-7\%$ when trained partially on Tomita languages. We also observe a similar trend with CFGs such as Dyck grammar, where O2RNN exhibits stable performance.  Furthermore, we also showed that the equivalence query-based approach, $L^*$, exhibits a significant degree of instability in DFA extraction. Moreover, we also show even on a partially trained network, the quantization-based approach constantly outperforms $L^*$, indicating a promising direction on the importance of the quantization-based DFA extraction approach coupled with tensor connections.

\bibliography{arxiv_ref}
\bibliographystyle{plainnat}

\newpage
\appendix
\onecolumn
\color{black}
\section{Tomita Grammars}
\begin{table}[H]
\centering

\begin{tabular}{|c|c|}
    
    \hline
    \# & Definition \\
    \hline
    1 & $a^*$ \\
    2 & $(ab)^*$ \\
    3 & Odd ${\llbracket .... \rrbracket}_a$ of $a$'s must be followed by even ${\llbracket .... \rrbracket}_b$ of $b$'s \\
    4 & All strings without the trigram $aaa$ \\
    5 & Strings $I$ where ${\llbracket .... \rrbracket}_a(I)$ and ${\llbracket .... \rrbracket}_b(I)$ are even \\
    6 & Strings $I$ where ${\llbracket .... \rrbracket}_a(I)$ $\equiv_3$$ \llbracket .... \rrbracket_b(I)$ \\
    7 & $b^*a^*b^*a^*$ \\
    \hline
\end{tabular}

\caption{\color{black} Definitions of the Tomita languages. Let ${\llbracket .... \rrbracket}_{\sigma}(I)$ denote the number of occurrences of symbol $\sigma$ in string $I$. Let $\equiv_3$ denote equivalence mod $3$ (\cite{mali2023computational}).}
\label{tab:tomita}
\end{table}

\section{$L^*$ Algorithm}
\label{app:lstar}
\color{black}
$\mathbf{L^{*}}$ algorithm, proposed by \cite{angluin1987learning} learns an unknown regular set $U$ over a fixed known finite alphabet $A$ from a minimally adequate Teacher $T$. $U$ is represented as a DFA. A minimally adequate Teacher with knowledge of regular set $S$ is any system that can answer two types of queries:
\begin{enumerate}
	\item Membership Query: For a given Teacher $T$ and a membership query $Q_M$ for string $s$ the teacher responds as :
	$$
		Q_M(T, s) = \begin{cases}
				1  \quad s \in S \\
				0	\quad \text{otherwise} 
              \end{cases}
	$$
	\item Equivalence Query: The Teacher $T$ responds to an equivalence query $Q_C$ as:
 $$
    Q_C(T, U) = \begin{cases}
                1 \quad \text{if}\ S \subseteq U \text{and}\ U \subseteq S \\
                t \quad t \in \ S \cup U - S \cap U
    \end{cases}
 $$
\end{enumerate}
$L^*$ maintains an observation table as a Hankel Matrix ($H(p.e) \rightarrow \{0, 1\}$) where $p \in P \cup P.A$, where $P$ is a finite prefix closed set. $b \in E$, where $E$ is a finite suffix closed set. $L^{*}$ performs a sequence of $Q_M$ and $Q_C$ with the given teacher $T$ and concludes when $H$ is closed and consistent. $H$ is closed iff $\forall t \in P.A \ \exists \ p' \in P $ s.t. row($p'$) = row($t$) and $H$ is consistent iff for $p_1, p_2 \in P $ and row($p_1$) = row($p_2$), then $\forall a \in A$ row($p_1.a$) = row($p_2.a$). Here $(.)$ is the string concatenation operator. A DFA $(Q, \Sigma, \delta, q_0, F)$ is then constructed from $H$ with $\Sigma = A$, $Q = \{ row(p) : \forall \ p \in P \}$, $q_0 = row(\epsilon)$, $F = row(p): p \in P \ \text{and}\ H(p) = 1$, and $\delta(row(p), a) = row(p.a)$

In the absence of a minimally adequate teacher $T$, a stochastic Oracle $O$ capable of answering $Q_M$ and $Q_C$ can also be used.  We follow the approach of \cite{weiss2018extracting} and use the RNNs as oracle to extract DFAs for each of the $3200$ models. One observation we made here is that the extraction heavily depends on initial counter-examples. This does not prove good for Dyck grammars especially.

\color{black}
\section{Training Details}
\label{sec:appendix_training}

The weights of all 4 RNNs are initialized using fan-in/fan-out initialization; biases are initialized to $1$. The batch size is set to $2048$, and all networks are optimized using stochastic gradient descent with an initial learning rate of $0.01$ for a maximum of $15000$ iterations with a single-cycle learning rate scheduler \cite{smith2019super}. We also use early stopping with patience criteria of $1000$ iterations on the validation set. All models are trained using binary-cross entropy.  We use grid search on learning rate in the range $[1, 1e-5]$to find the optimal set of hyper-parameters. We also introduce a data augmentation approach that helps networks with fewer parameters to converge on difficult grammars such as Tomita-$3$ and Tomita-$6$. At each step, we classify partial string and calculate the loss at that particular timestep.

Prior works focused on recognizing grammatical sequences \cite{bhattamishra2020ability, mali2021investigating, suzgun2019lstm} have shown that neural networks with few parameters can effectively learn Tomita and Dyck languages. However, works focused on rule extraction \cite{weiss2018extracting} have used large networks with hidden sizes ($>50$) and layers ($>1$). This doesn't align with theoretical results, which state that a single-layer network can recognize regular grammars \cite{Merrill_2019, mali2023computational}. In some cases, it is difficult to evaluate the memorization effect vs the generalization capability of the model if the number of parameters is more than the samples. Thus, to better understand the true generalization capability of models, we train our networks on a small range of hidden state sizes. For Tomita grammars, we use hidden sizes, which are multiples of the number of states in the ground truth DFA. Table \ref{tab:tomita_states} provides the number of states of minimal DFA designed to recognize 7 Tomita grammars. For Dyck-$k$ languages, we use multiples of $k$ as hidden size. Furthermore, we also compare the stability of extraction of DFAs on a hidden state size multiple of $2$, but we also train our models on multiples of $1$, $3$, $4$, and $5$. The results of extended multiples are covered in the additional results section.  In Table \ref{tab:num_params}, we report the number of parameters for the smallest and the largest models used to recognize each language. In our experiments, the smallest model had only $\mathbf{21}$ parameters, whereas the largest model had $\mathbf{3126}$. 

\begin{table}[h]
\begin{tabular}{llllllll}
\hline
         & \multicolumn{7}{c}{Tomita Grammars} \\
         & 1   & 2   & 3   & 4  & 5  & 6  & 7  \\ \hline
\#states & 2   & 3   & 5   & 4  & 4  & 3  & 5  \\ \hline
\end{tabular}
\centering
\caption{Minimal number of states in the ground truth DFA of Tomita grammars}
\label{tab:tomita_states}
\end{table}

\textbf{Computational budget:} As notified in prior sections we train $4$ types of RNN models for $7$ Tomita grammars and $2$ dyck grammars, with $5$ configurations of hidden state sizes, across $10$ initial seeds. Additionally, we have partially trained RNNs with all the above configurations for 7 Tomita grammars, resulting in a total $3200$ neural networks. Each network takes, on average, approximately $30$ mins to train on a single Nvidia $2080$ti GPU. Thus, we spend a total of $1600$ GPU hours for training. Testing and DFA extraction required additional GPU hours.

\section{Additional Results}
\label{sec:add_result}
  
\begin{table}[h]
\begin{tabular}{llll}
\hline
           & \begin{tabular}[c]{@{}l@{}}String Length\\ {[}min - max{]}\end{tabular} & \begin{tabular}[c]{@{}l@{}}Tomita \\ {[}1-7{]}\end{tabular} & \begin{tabular}[c]{@{}l@{}}Dyck\\  {[}2,3,6,8{]}\end{tabular} \\ \hline
Train      & 2-50                                                                    & 10000                                                       & 20000                                                         \\
Val        & 2-50                                                                    & 2000                                                        & 4000                                                          \\
Test Bin 0 & 2-50                                                                    & 2000                                                        & 4000                                                          \\
Test Bin 1 & 51-100                                                                  & 2000                                                        & 4000  \\ \hline
\end{tabular}
\centering
\caption{Number of samples for each grammar in the datasets}
\label{tab:num_samples}
\end{table}

\begin{table}[]
\begin{tabular}{llllll}
\hline
grammar                                        & \begin{tabular}[c]{@{}l@{}}state\\ dim\end{tabular} & lstm & gru  & o2rnn & mirnn \\ \hline
\multicolumn{1}{c}{\multirow{2}{*}{tomita\_1}} & 2                                                   & 67   & 51   & 21    & 23    \\
\multicolumn{1}{c}{}                           & 10                                                  & 651  & 491  & 421   & 191   \\ \hline
\multirow{2}{*}{tomita\_2}                     & 3                                                   & 112  & 85   & 43    & 37    \\
                                               & 15                                                  & 1276 & 961  & 931   & 361   \\ \hline
\multirow{2}{*}{tomita\_3}                     & 5                                                   & 226  & 171  & 111   & 71    \\
                                               & 25                                                  & 3126 & 2351 & 2551  & 851   \\ \hline
\multirow{2}{*}{tomita\_4}                     & 4                                                   & 165  & 125  & 73    & 53    \\
                                               & 20                                                  & 2101 & 1581 & 1641  & 581   \\ \hline
\multirow{2}{*}{tomita\_5}                     & 4                                                   & 165  & 125  & 73    & 53    \\
                                               & 20                                                  & 2101 & 1581 & 1641  & 581   \\ \hline
\multirow{2}{*}{tomita\_6}                     & 3                                                   & 112  & 85   & 43    & 37    \\
                                               & 15                                                  & 1276 & 961  & 931   & 361   \\ \hline
\multirow{2}{*}{tomita\_7}                     & 5                                                   & 226  & 171  & 111   & 71    \\
                                               & 25                                                  & 3126 & 2351 & 2551  & 851   \\ \hline
\multirow{2}{*}{dyck\_2}                       & 2                                                   & 91   & 69   & 33    & 29    \\
                                               & 10                                                  & 771  & 581  & 721   & 221   \\ \hline
\multirow{2}{*}{dyck\_3}                       & 3                                                   & 172  & 130  & 88    & 52    \\
                                               & 15                                                  & 1576 & 1186 & 2056  & 436   \\ \hline
\multirow{2}{*}{dyck\_6}                       & 5                                                   & 559  & 421  & 553   & 157   \\
                                               & 30                                                  & 5671 & 4261 & 13561 & 1501  \\ \hline
\multirow{2}{*}{dyck\_8}                       & 8                                                   & 937  & 705  & 1233  & 257   \\
                                               & 40                                                  & 9801 & 7361 & 30481 & 2561  \\ \hline
\end{tabular}
\centering
\caption{Number of parameters in RNNs for smallest and largest hidden state size used for each grammar}
\label{tab:num_params}
\end{table}

\begin{figure*}[!hbt]
\includegraphics[width=0.8\textwidth]{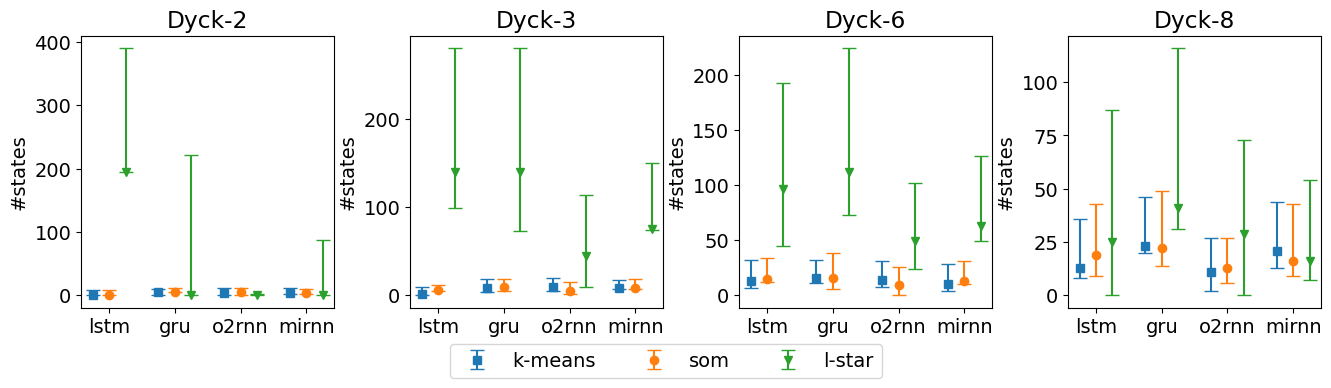}
\centering
\caption{Min, max and mode of the number of states extracted from $1^{st}$ and $2^{nd}$ order RNNs on Dyck grammars.}

\label{fig:dyck_comparison_states_std}
\end{figure*}

\begin{figure*}[!htb]
\includegraphics[width=0.8\textwidth]{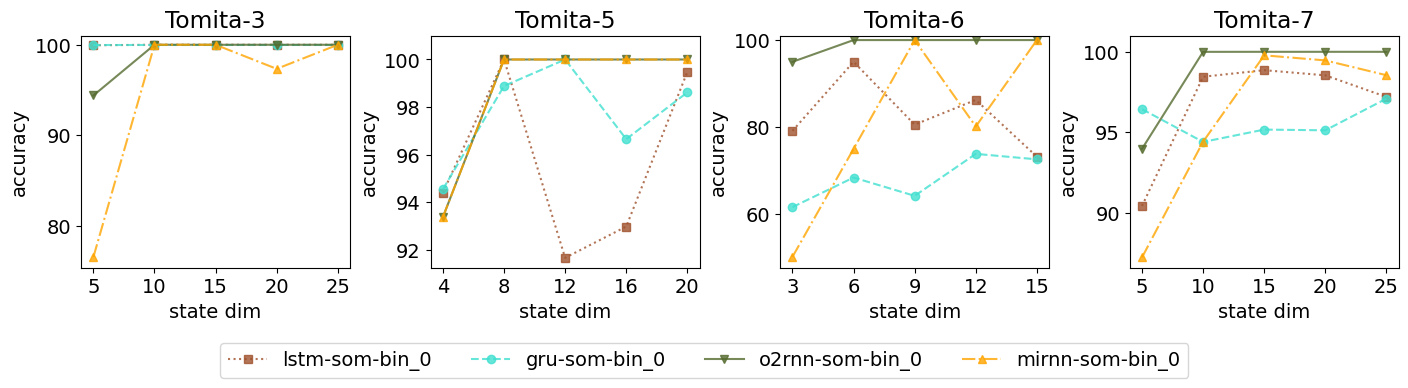}
\centering
\caption{Mean accuracy of DFAs extracted from various RNNs using quantization method (clustering by SOM).}
\label{fig:tomita_dfa_accuracy}
\end{figure*}

\begin{figure*}[!htb]
\includegraphics[width=0.8\textwidth]{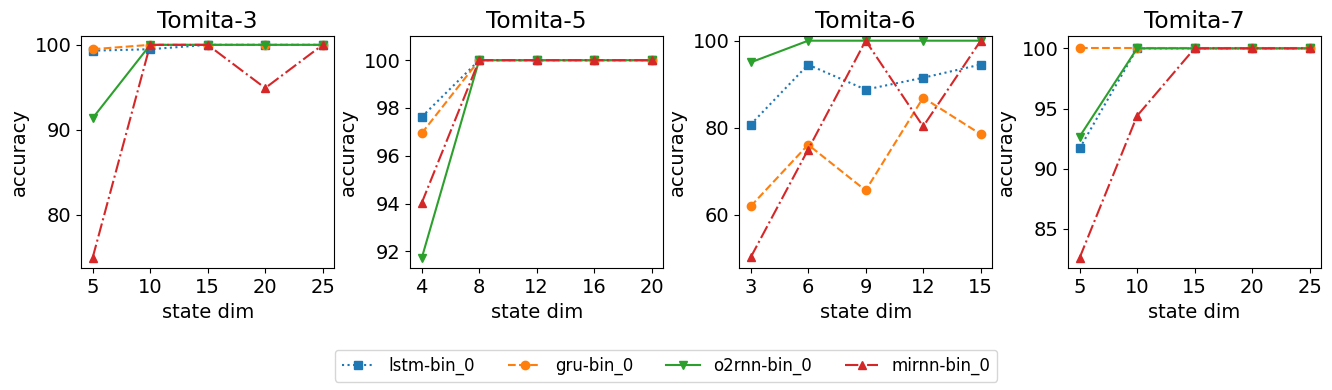}
\centering
\caption{Mean accuracy of DFAs extracted from various RNNs using equivalence query method ($L^{*}$)}
\label{fig:tomita_lstar_accuracy}
\end{figure*}

\begin{figure*}[!htb]
\includegraphics[width=0.8\textwidth]{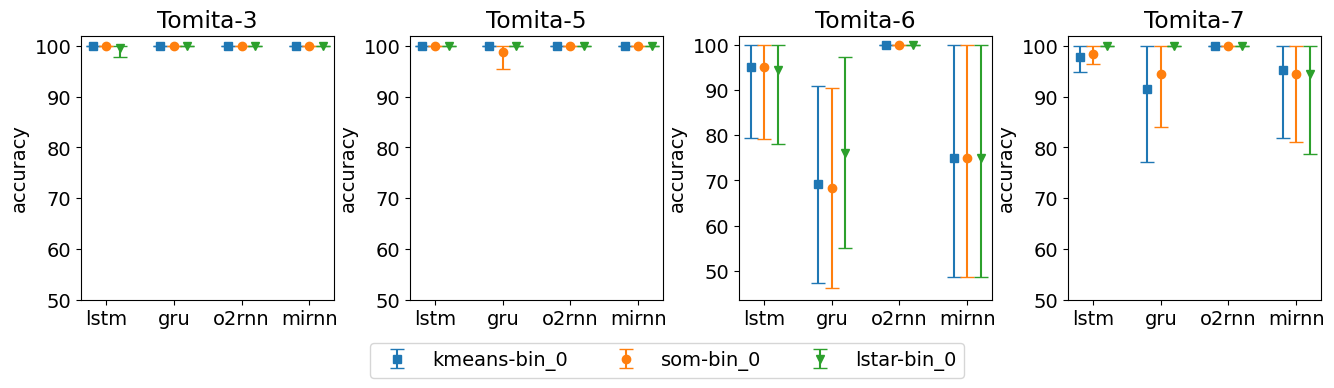}
\centering
\caption{Mean and standard deviation of accuracy of DFAs extracted from $1^{st}$ and $2^{nd}$ order RNNs trained on Tomita grammars over $10$ seeds}
\label{fig:tomita_dfa_accuracy_comparison}
\end{figure*}

\begin{figure*}[!htb]
\includegraphics[width=0.8\textwidth]{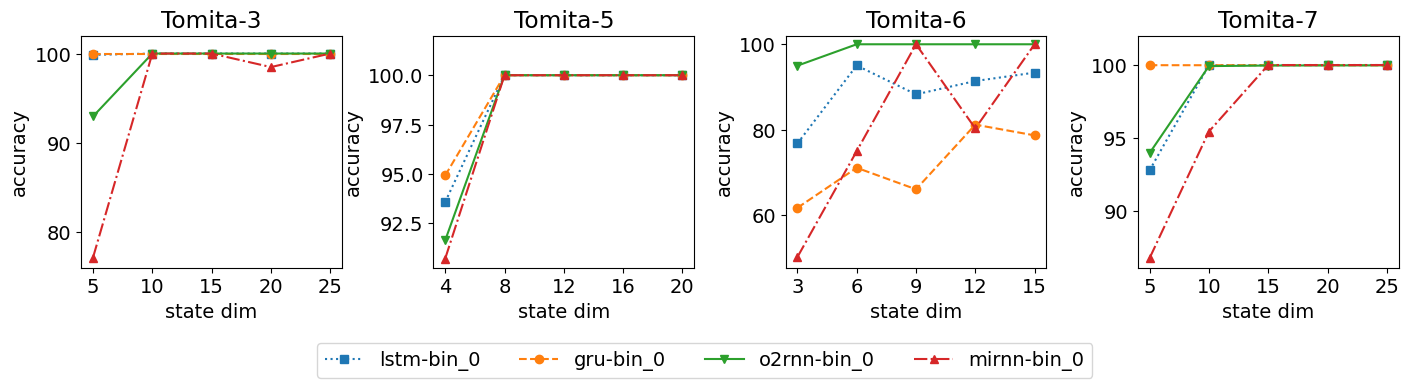}
\centering
\caption{Mean Accuracy of $1^{st}$ and $2^{nd}$ order Recurrent Neural Networks on Tomita $3$, $5$, $6$ and $7$}
\label{fig:tomita_rnn_mean}
\end{figure*}

\begin{figure*}[!htb]
\includegraphics[width=0.6\textwidth]{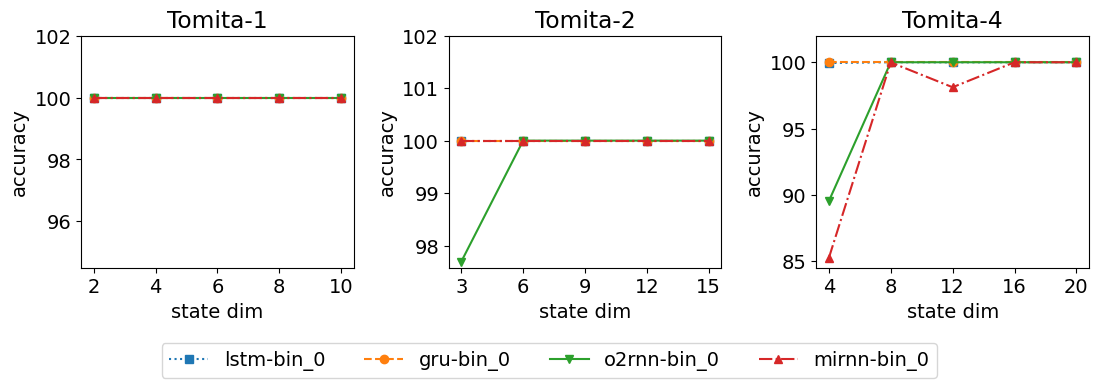}
\centering
\caption{Mean Accuracy of $1^{st}$ and $2^{nd}$ order Recurrent Neural Networks on Tomita $1$, $2$ and $4$}
\label{fig:tomita_124_rnn_mean}
\end{figure*}

\begin{figure*}[!htb]
\includegraphics[width=0.8\textwidth]{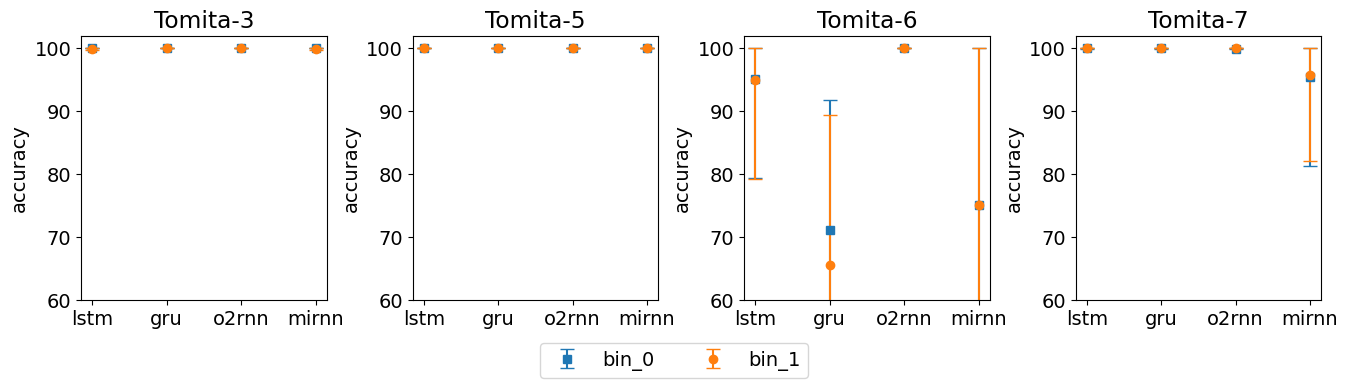}
\centering
\caption{Mean and Standard Deviation of $1^{st}$ and $2^{nd}$ order Recurrent Neural Networks trained on $3$, $5$, $6$ and $7$.}
\label{fig:tomita_rnn_std}
\end{figure*}

\begin{figure*}[!htb]
\includegraphics[width=0.6\textwidth]{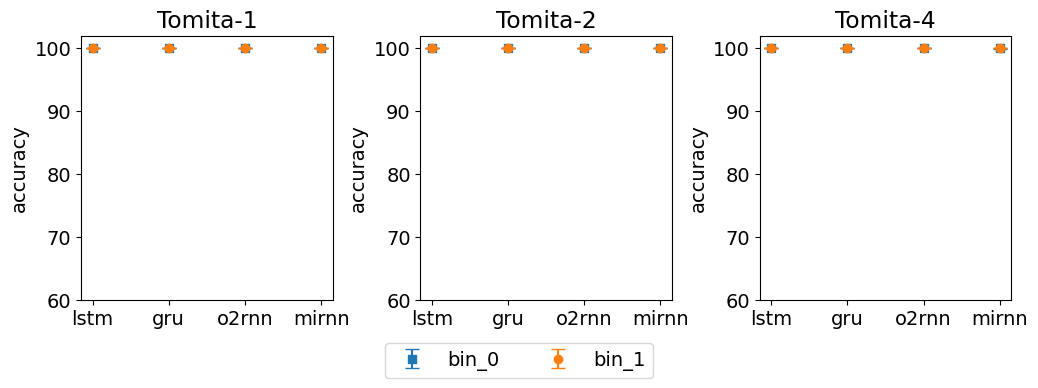}
\centering
\caption{Mean and Standard Deviation of $1^{st}$ and $2^{nd}$ order Recurrent Neural Networks trained on Tomita $1$, $2$ and $4$.}
\label{fig:tomita_124_rnn_std}
\end{figure*}

\begin{figure*}[!htb]
\includegraphics[width=0.55\textwidth]{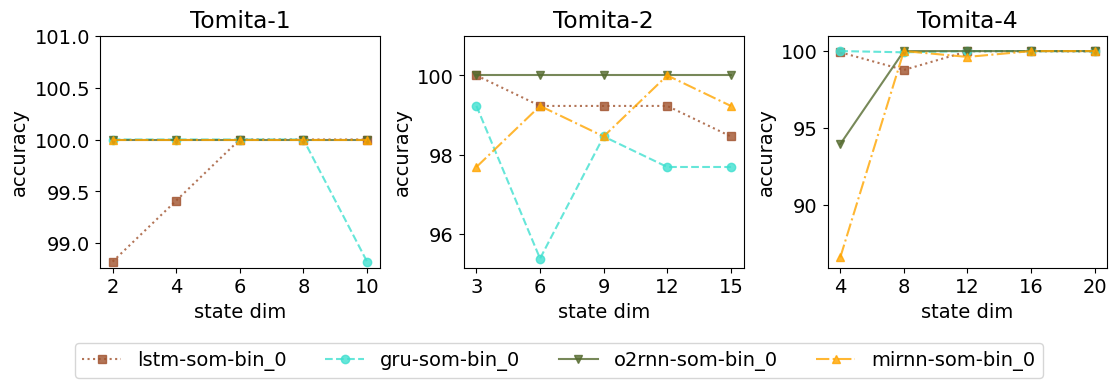}
\centering
\caption{Mean accuracy of DFA extracted from $1^{st}$ and $2^{nd}$ order Recurrent Neural Networks trained on Tomita $1$, $2$ and $4$ by quantization methods}
\label{fig:tomita_124_rnn_accuracy}
\end{figure*}

\begin{figure*}[!htb]
\includegraphics[width=0.55\textwidth]{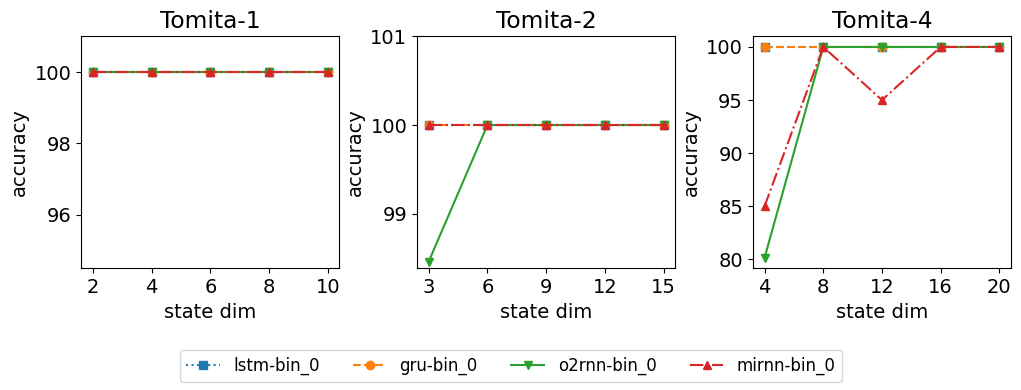}
\centering
\caption{Mean accuracy of DFA extracted from $1^{st}$ and $2^{nd}$ order Recurrent Neural Networks trained on Tomita $1$, $2$ and $4$ by $L^{*}$}
\label{fig:tomita_124_rnn_dfa_accuracy}
\end{figure*}

\begin{figure*}[!htb]
\includegraphics[width=0.55\textwidth]{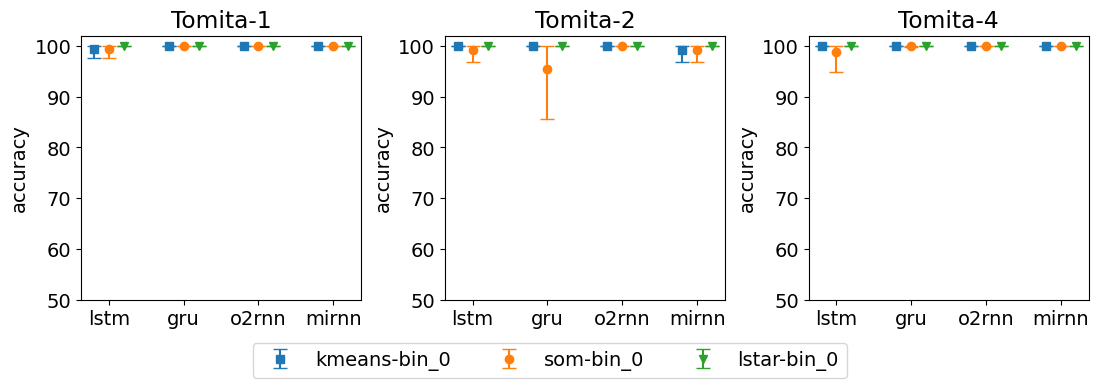}
\centering
\caption{Mean and Standard Deviation of DFA extracted from of $1^{st}$ and $2^{nd}$ order Recurrent Neural Networks trained on Tomita $1$, $2$ and $4$.}
\label{fig:tomita_124_rnn_dfa_std}
\end{figure*}

\begin{figure*}[!htb]
\includegraphics[width=0.55\textwidth]{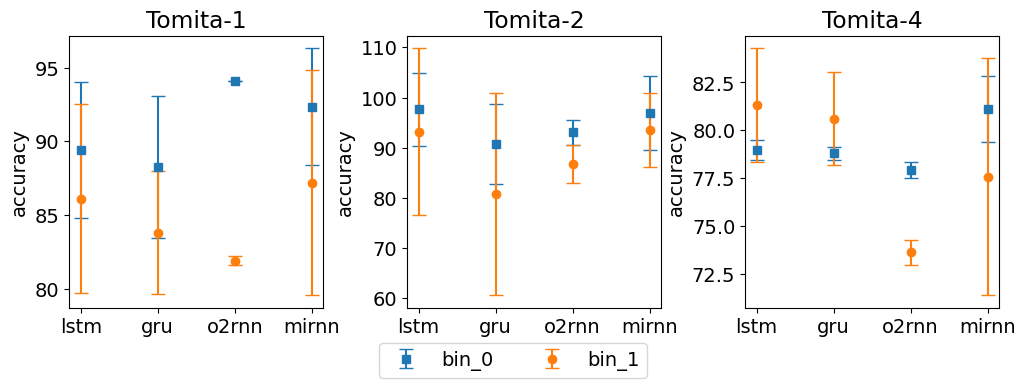}
\centering
\caption{Mean and Standard Deviation of $1^{st}$ and $2^{nd}$ order Recurrent Neural Networks partially trained on Tomita $1$, $2$ and $4$.}
\label{fig:tomita_124_rnn_partial_std}
\end{figure*}

\begin{figure*}[!htb]
\includegraphics[width=0.55\textwidth]{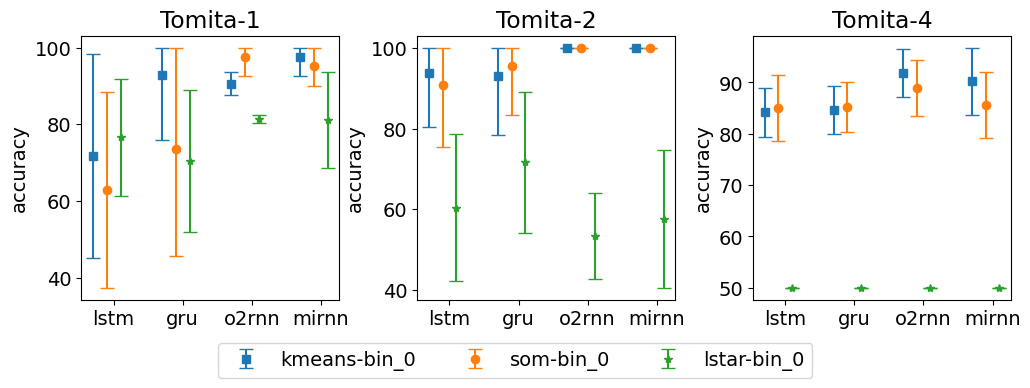}
\centering
\caption{Mean and Standard Deviation of DFA extracted from $1^{st}$ and $2^{nd}$ order Recurrent Neural Networks partially trained on Tomita $1$, $2$ and $4$.}
\label{fig:tomita_124_rnn_dfa_partial_std}
\end{figure*}


\begin{figure*}[!htb]
\includegraphics[width=0.8\textwidth]{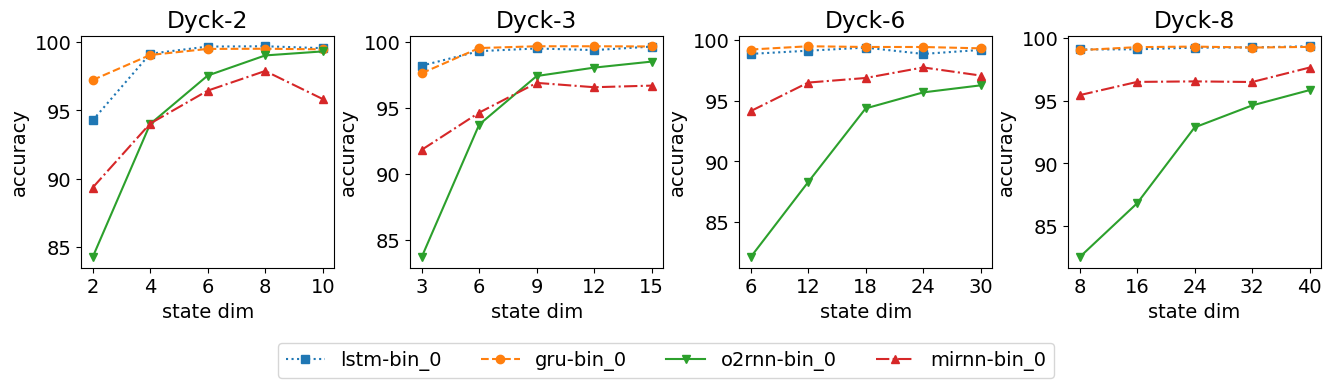}
\centering
\caption{Mean Accuracy of $1^{st}$ and $2^{nd}$ order Recurrent Neural Networks on Dyck-2, 3, 6 and 8 languages}
\label{fig:dyck_rnn_mean}
\end{figure*}
\begin{figure*}[h]
\includegraphics[width=0.8\textwidth]{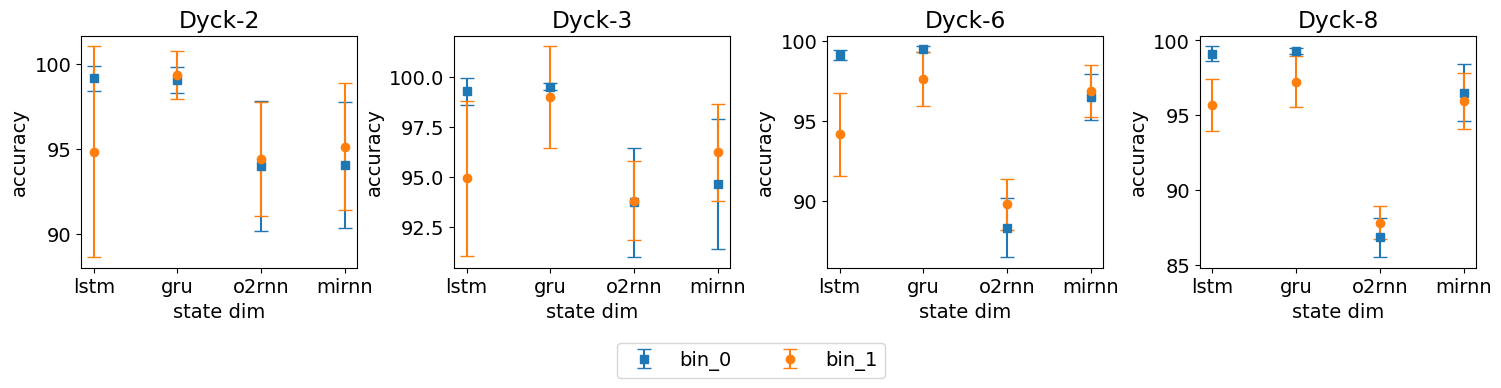}
\centering
\caption{Mean Accuracy and Standard Deviation of $1^{st}$ and $2^{nd}$ order Recurrent Neural Networks trained on Dyck grammars}
\label{fig:dyck_rnn_std}
\end{figure*}

\begin{figure*}[!htb]
\includegraphics[width=0.8\textwidth]{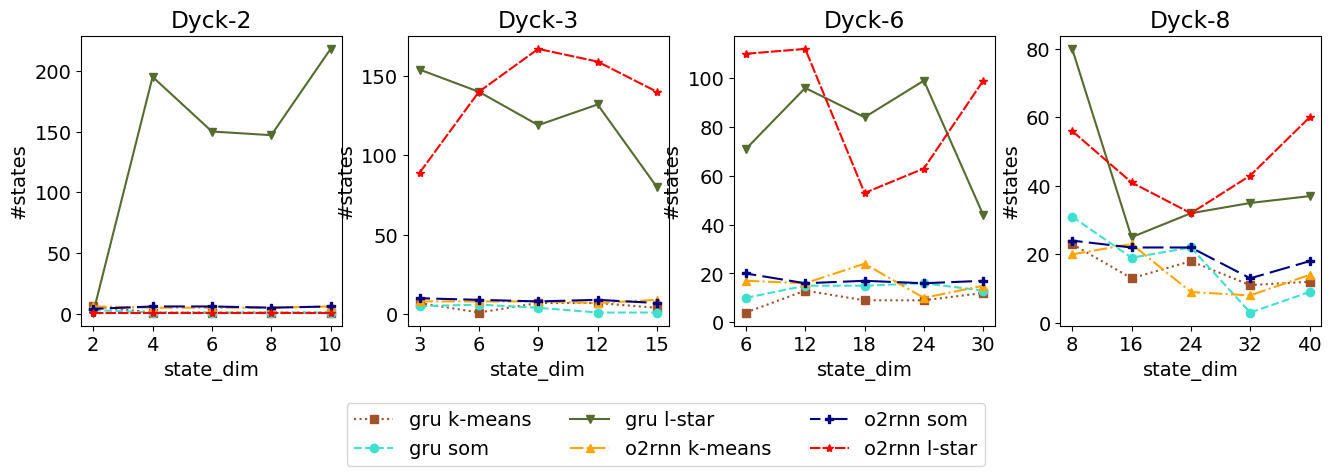}
\centering
\caption{Mode of the number of states extracted from $1^{st}$ and $2^{nd}$ order RNNs on Dyck grammars across $10$ seeds.}

\label{fig:dyck_comparison_states_mode}
\end{figure*}

\end{document}